\documentclass{article}
\usepackage{spconf, amsmath, graphicx}

\usepackage{booktabs}
\usepackage{amssymb} 
\usepackage[capitalize]{cleveref}  
\crefname{section}{Sec.}{Secs.}
\Crefname{section}{Section}{Sections}
\crefname{table}{Tab.}{Tabs.}
\Crefname{table}{Table}{Tables}
\usepackage{color}
\usepackage{kotex}
\usepackage{caption}
\usepackage{subcaption}
\usepackage{multirow}

\DeclareMathOperator*{\argmin}{arg\,min}

\title{Exploration into Translation-Equivariant Image Quantization}
\name{Woncheol Shin$^{\dagger}$ \quad Gyubok Lee$^{\dagger}$ \quad Jiyoung Lee$^{\dagger}$ \quad Eunyi Lyou$^{\ddagger}$ \quad Joonseok Lee$^{\ddagger \star}$ \quad Edward Choi$^{\dagger}$
\thanks{This work was supported by the KAIST-NAVER Hyper-Creative AI Center and the Institute of Information \& communications Technology Planning \& Evaluation (IITP) grant (No.2019-0-00075, No.2022-0-00984) and National Research Foundation of Korea (NRF) grant (NRF-2020H1D3A2A03100945, NRF-2021H1D3A2A03038607) funded by the Korea government (MSIT).}
}

\address{
$^{\dagger}$ Graduate School of AI, KAIST, Daejeon, South Korea \\
$^{\ddagger}$ Graduate School of Data Science, Seoul National University, Seoul, South Korea \\
$^{\star}$ Google Research, Mountain View, California, USA
}

\begin{document}
%
\maketitle
\begin{abstract}
This is an exploratory study that discovers the current image quantization (vector quantization) do not satisfy translation equivariance in the quantized space due to aliasing.
Instead of focusing on anti-aliasing, we propose a simple yet effective way to achieve translation-equivariant image quantization by enforcing orthogonality among the codebook embeddings.
To explore the advantages of translation-equivariant image quantization, we conduct three proof-of-concept experiments with a carefully controlled dataset:
(1) text-to-image generation, where the quantized image indices are the target to predict,
(2) image-to-text generation, where the quantized image indices are given as a condition,
(3) using a smaller training set to analyze sample efficiency.
From the strictly controlled experiments, we empirically verify that the translation-equivariant image quantizer improves not only sample efficiency but also the accuracy over VQGAN up to +11.9\% in text-to-image generation and +3.9\% in image-to-text generation.
\end{abstract}
\section{Introduction}
\label{sec:intro}

\begin{figure}[t]
   \centering
   \includegraphics[width=0.9\linewidth]{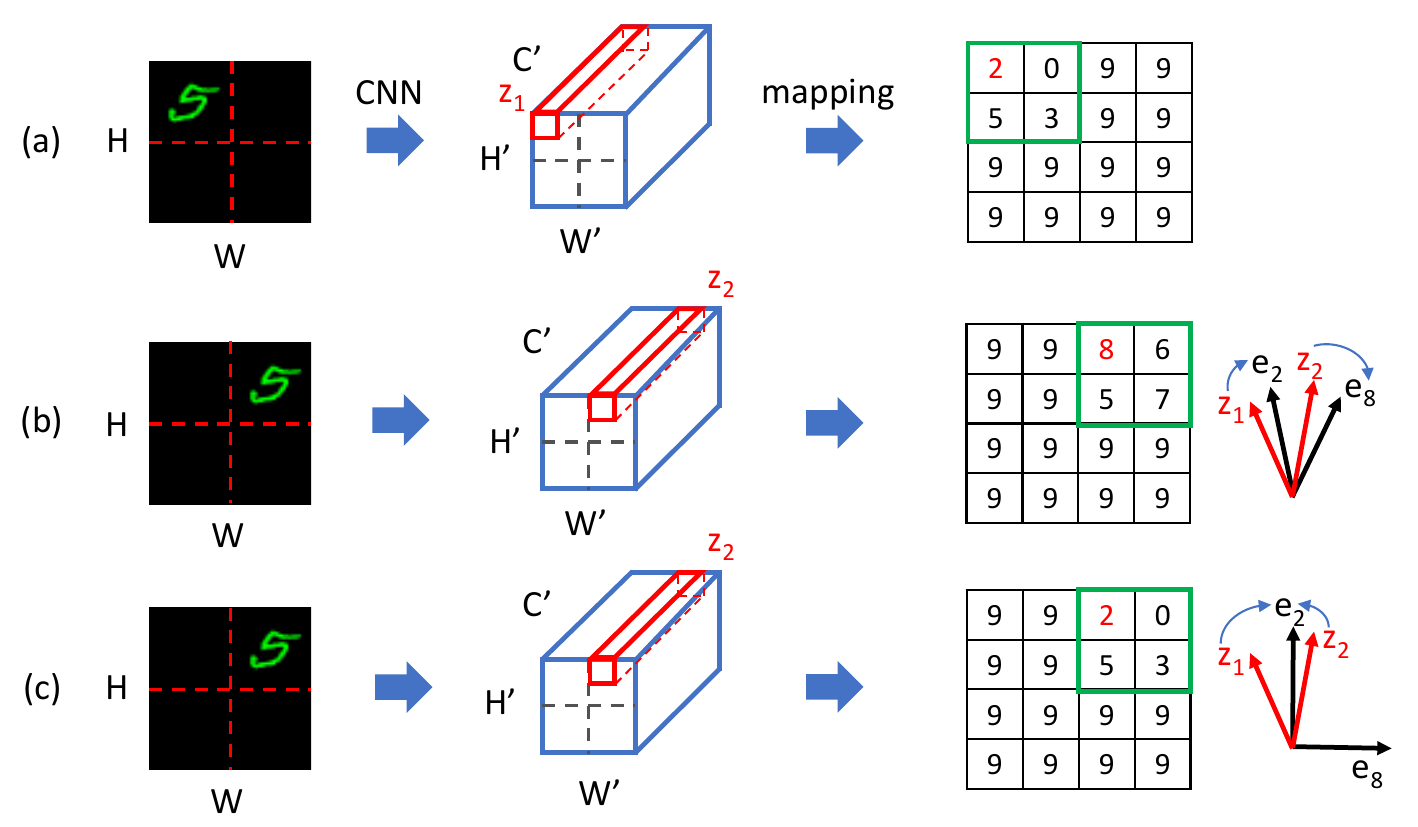}
   \vspace{-1mm}
   \caption{(a) Image quantization before translation of the top-left 5; (b) Broken translation equivariance in the quantized space after the translation of 5; (c) Perfect translation equivariance in the quantized space with our orthogonal codebook embeddings. $z_1, z_2$ are the corresponding features in terms of shifted location. By enforcing orthogonality in the quantized space as in (c), a slight deviation due to aliasing can be ignored, and the same codebook index (\textit{e.g.} 2) is given to $z_1, z_2$.}
    \label{fig:idea}
\end{figure}
\vspace{-1mm}

Vector quantization~\cite{oord2017neural} has gained popularity in multimodal learning problems such as text-to-image generation. In particular, several works~\cite{ramesh2021zero, ding2021cogview, yu2022scaling} demonstrated impressive text-conditioned image generation, only using image quantization and Transformer~\cite{vaswani2017attention}.
Methods that utilize vector quantization in image downstream tasks usually take a two-stage approach: first learning to quantize images, then solving a downstream task with the quantized representation.
In the first step, an image quantizer (\textit{e.g.}, VQVAE~\cite{oord2017neural} or VQGAN~\cite{esser2021taming}) encodes an image as a sequence of codebook indices using codebook embeddings (Stage 1). Then, the resulting indices are given as input to downstream image modeling tasks (Stage 2). 
By modeling images in this fashion, the input indices of the downstream task can be treated in the same manner as token indices in text. This discretization of images allows us to model a relatively shorter sequence of image representations, and to even jointly handle textual information more naturally.

Although the idea of representing an image with quantized indices has been appreciated, the representations learned by existing methods turn out to be far from ideal yet. Specifically, we focus on \emph{translation equivariance}, a property for an image quantizer to represent semantically same objects with the same indices regardless of its location within the image. A toy example in \Cref{fig:idea} (b) indicates, however, that the state-of-the-art image quantization method, VQGAN~\cite{esser2021taming}, breaks this property even for the most trivial setting (locating the exactly same image patch at the pixel level).

Referring to the literature, the cause of the broken translation equivariance is due to aliasing, caused by downsampling operations (\textit{e.g.}, strided convolution, maxpooling)~\cite{simoncelli1992shiftable,zhang2019making}.  In other words, the following can hold:
\vspace{-1mm}
\begin{equation}
  z_2 = z_1 + \epsilon,
\vspace{-1mm}
\end{equation}
where $\epsilon$ is a noise caused by aliasing, and $z_1$ and $z_2$ are the corresponding features in terms of the shifted location (\Cref{fig:idea}). 
In an effort to reduce $\epsilon$, one of the anti-aliasing methods in both signal processing and CNNs is to apply a low-pass filter during downsampling~\cite{oppenheim1999discrete,gonzalez2009digital,zhang2019making,karras2021alias}. 
In deep neural networks, it is challenging to enforce $\epsilon$ to be completely zero in the feature map space, as complete anti-aliasing still requires careful manipulation of filters.
However, when it comes to image quantization, where images are represented as codebook indices, $\epsilon$ no longer needs to be zero as long as the features are mapped to the same codebook index. 
This implies that translation equivariance in the `quantized space' can be achieved by a different approach than enforcing translation equivariance in the `feature map space.'

In this paper, we propose a simple yet effective way to achieve translation-equivariant image quantization by enforcing orthogonality among the codebook embeddings.
Then, with carefully controlled datasets, we study how it affects the downstream image modeling tasks.
For the Stage 2, we use the following three settings:
(1) text-to-image generation, where the quantized image indices are the target to predict,
(2) image-to-text generation, where the quantized image indices are given as a condition,
(3) using a smaller training set to analyze sample efficiency.
We report advantages and limitations of translation-equivariant image quantization in all three settings from thorough analysis of the experimental results.

Our contributions include:
(1) To the best of our knowledge, this is the first work to explore the problem of \emph{translation equivariance in the quantized image space}; 
(2) Instead of focusing on anti-aliasing, we take a direct approach to achieve translation equivariance in the quantized space by regularizing orthogonality in the codebook embedding vectors; and 
(3) We show that a translation-equivariant image quantizer improves not only sample efficiency but also the accuracy of text-to-image and image-to-text generation on text-augmented MNIST~\cite{mnist} by up to +11.9\% and +3.9\%, respectively. We discuss insights discovered from our analysis on the behavior of the quantized representations.


\section{Related Work}
\vspace{-1mm}
\label{sec:related_work}

\subsection{Image Quantization}
\vspace{-1mm}

Image quantization or vector quantization of images is an efficient encoding method that represents images in the discrete latent space via an encoder and decoder.
Oord et al. \cite{oord2017neural} first applied this idea to the generative tasks, which can fully leverage the power of both encoder and decoder.
In this manner, downstream models (generators) only need to handle a much shorter sequence of image tokens, compared to handling each pixel.
Thanks to the other advantage of jointly handling image and text tokens more naturally, this idea quickly spread to multimodal tasks such as text-to-image generation.
For example, DALL-E~\cite{ramesh2021zero} and CogView~\cite{ding2021cogview} receive both the text and image tokens as a single stream, similar to GPT~\cite{brown2020language}, and calculate self-attention.
Parti~\cite{yu2022scaling} adopts an encoder-decoder Transformer~\cite{vaswani2017attention} structure, where the encoder takes text tokens as inputs and the decoder autoregressively predicts discrete image tokens.
These models demonstrated the effectiveness of the quantization in the text-image multimodal problem.

Esser et al. \cite{esser2021taming} proposed VQGAN, an effective extension of VQVAE~\cite{oord2017neural} by introducing an adversarial loss and perceptual loss.
The addition of two objectives provides image representations that result in a sharper and detailed reconstruction of images.
More recently, Yu et al. \cite{yu2021vector} further boosted the efficiency and reconstruction quality of image quantizers, replacing CNNs with the ViT architecture~\cite{dosovitskiy2020image}.
Since Convolutional Neural Network (CNN) is one of the most widely used and well-established networks, we utilize a CNN-based quantizer, VQGAN, as our base image quantizer in this work.

\vspace{-1mm}
\subsection{Translation Invariance and Equivariance}
\vspace{-1mm}

Translation invariance requires the output unchanged by the shifts in the input, while translation equivariance is a mapping which, when the input is shifted, leads to a shifted output. 
In other words, translation invariance is about the final representation after Global Average Pooling (GAP) in CNNs, and translation equivariance is about the feature map before GAP.
A fundamental approach to handle translation invariance and equivariance is anti-aliasing. Simoncelli et al. \cite{simoncelli1992shiftable} first formalized `shiftability' and related it to aliasing. Since then, careful calibration of sampling rate according to the Nyquist sampling theorem~\cite{nyquist1928certain} or applying a low-pass filter has been a natural choice to avoid aliasing in downsampling. 

It is only recently that deep learning has started to explore translation invariance and equivariance~\cite{azulay2018deep,zhang2019making,zou2020delving,vasconcelos2021impact}. 
Zhang \cite{zhang2019making} applied a low pass filter between a stride-one operator and naive subsampling to improve translation equivariance in the latent feature map space, but Azulay and Weiss \cite{azulay2018deep} pointed out that nonlinearity such as ReLU still hinders anti-aliasing even with low-pass filtering. Karras et al. \cite{karras2021alias} proposed an ideal sampling method and nonlinearity to avoid aliasing in the image generation task. In a similar motivation to theirs, we investigate translation equivariance in the generation task but pay special attention to the vector-quantized space and multimodal learning problem.


\section{Method}
\label{sec:method}


\subsection{Translation Equivariance in Quantized Space}
\label{ssec:te}

Let $\mathbf{x} \in \mathbb{R}^{H \times W \times 3}$ be an image of size $H \times W$ with RGB channels. A CNN $\mathcal{F}$ takes $\mathbf{x}$ as input, and produces a feature map $\mathcal{F}(\mathbf{x}) \in \mathbb{R}^{H' \times W' \times C'}$. 
Let $Q$ be a quantization operation, and $Q(\mathcal{F}(\mathbf{x})) = \mathcal{F}^q(\mathbf{x}) \in \mathbb{R}^{H' \times W' \times C'}$ be the quantized feature map. $\mathcal{F}$ is called \emph{translation equivariant in the quantized space} if
\begin{equation} 
  \mathcal{F}^q(T_1(\mathbf{x})) = T_2(\mathcal{F}^q(\mathbf{x}))
\end{equation}
where $T_1$ is a translation operation in the image space and $T_2$ is the translation operation in the quantized space corresponding to $T_1$.
In other words, it is enough for the translation equivariance in the quantized space to have the $T_2$ relationship between the quantized code indices of $\mathcal{F}(T_1(\mathbf{x}))$ and $\mathcal{F}(\mathbf{x})$, even if $\mathcal{F}(T_1(\mathbf{x})) \neq T_2(\mathcal{F}(\mathbf{x}))$.
Note that we can safely focus only on the relationship between the code indices, because the decoder-only Transformer~\cite{vaswani2017attention} in the following stage takes the input image in the form of code indices, not image feature maps or codebook embedding vectors.

\subsection{Image Quantizer (TE-VQGAN)}

VQGAN~\cite{esser2021taming} consists of a CNN-based encoder $\mathcal{F}$, a CNN-based decoder $\mathcal{G}$, and codebook embeddings $e \in \mathbb{R}^{C' \times K}$, where $K$ is the codebook size and $C'$ is the number of feature channels.
$\mathcal{F}$ gets an image $\mathbf{x} \in \mathbb{R}^{H \times W \times 3}$ and produces a feature map $\mathcal{F}(\mathbf{x}) \in \mathbb{R}^{H' \times W' \times C'}$. 
Let a fiber of the feature map whose spatial coordinate is $(h',w')$ be $\mathcal{F}(\mathbf{x})_{(h',w')} \in \mathbb{R}^{C'}$, where $(h',w') \in [1, H'] \times [1, W'] \text{ and } h',w' \in \mathbb{Z}$.
$\mathcal{F}(\mathbf{x})_{(h',w')}$ is assigned to the closest embedding vector $e_k \in \mathbb{R}^{C'}$ based on $L_2$ distance: 
\begin{equation}
\mathcal{F}^{q}(\mathbf{x})_{(h',w')} = e_k, \; k =\argmin_{j}\left\|\mathcal{F}(\mathbf{x})_{(h',w')}-e_{j}\right\|_{2}^{2}.
\end{equation}
The decoder $\mathcal{G}$ gets $\mathcal{F}^q(\mathbf{x})$ and produces the reconstructed image, $\hat{\mathbf{x}} = \mathcal{G}(\mathcal{F}^q(\mathbf{x})) \in \mathbb{R}^{H \times W \times 3}$. 
$\mathcal{F}, \mathcal{G}$, and $e$ are jointly trained by minimizing $\| \mathbf{x}-\hat{\mathbf{x}} \|_2$.

As mentioned in \Cref{sec:intro}, if we interpret alias as a noise due to downsampling, the following relationship may hold:
\begin{equation}
  \mathcal{F}(T_1(\mathbf{x}))_{(h',w')} = T_2(\mathcal{F}(\mathbf{x}))_{(h',w')} + \epsilon.
\end{equation}
Here, it is a non-trivial task to make $\epsilon$ zero. However, as mentioned in \Cref{ssec:te}, as long as we are only interested in the image code indices, we can focus only on assigning the same code $e_k$ to both $\mathcal{F}(T_1(\mathbf{x}))_{(h',w')}$ and $T_2(\mathcal{F}(\mathbf{x}))_{(h',w')}$ rather than trying to actually make $\epsilon = 0$.
Surprisingly, this can be achieved by simply enforcing the orthogonal structure in the quantized space; that is, by regularizing the embedding vectors $e_k$ to be orthogonal to each other.
The intuition behind this is well illustrated in \Cref{fig:idea}.
In the unregularized quantized space, feature maps $z_1$ and $z_2$ that slightly differ due to alias are mapped to respective embedding vectors $e_2$ and $e_8$, respectively.
In the orthogonal quantized space, however, every embedding vector is orthogonal to one another, and it is easier to ignore small noise due to alias when $z_1$ and $z_2$ are mapped to the same embedding vector $e_2$.

We add the following regularization term to the loss function of VQGAN to enforce orthogonality among the codebook embeddings:
\begin{equation}
  \label{eq:our_ortho_term}
  \mathcal{L}_\text{REG}(e) = \lambda \frac{1}{K^2} \left\|\ell_2(e)^{\top} \ell_2(e)-I_{K}\right\|_{F}^{2}
\end{equation}
where $I_{K} \in \mathbb{R}^{K \times K}$ and $\ell_2(e) \in \mathbb{R}^{C' \times K} $ denotes the identity matrix and an $L_2$-normalized code embedding along the first dimension, respectively, and $\| \cdot \|_{F}$ denotes the Frobenius norm.
We set $\lambda$ to 10 in all experiments provided in \Cref{sec:exp}.


\section{Experiments}
\label{sec:exp}

\begin{figure}
\centering
\begin{subfigure}{0.48\linewidth}
\includegraphics[width=\linewidth]{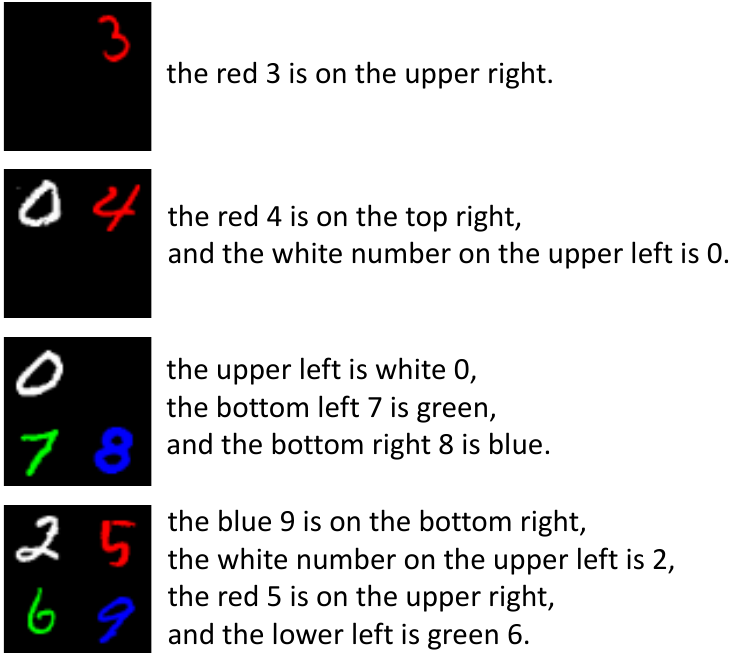}
\caption{Examples of the test set.}
\label{fig:mnist_stage2_testset}
\end{subfigure}
\hfill
\begin{subfigure}{0.48\linewidth}
  \includegraphics[width=\linewidth]{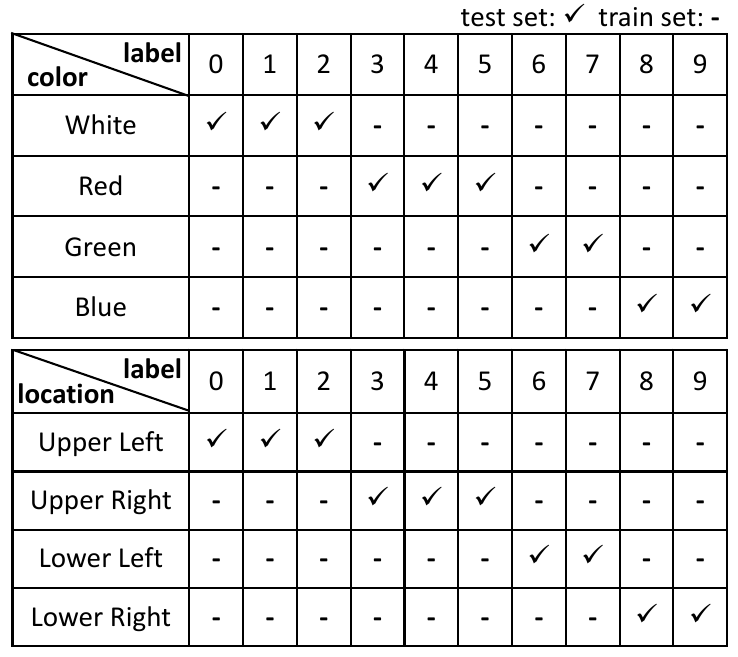}
  \caption{Constraints on the dataset.}
  \label{fig:dataset_constraint}
\end{subfigure}
\caption{Dataset examples and constraints.}
\label{fig:dataset}
\end{figure}
 
\noindent \textbf{Data Construction.}
To carefully explore the advantages of the translation-equivariant image quantizer, we created MNIST-based image-caption pairs, as shown in \Cref{fig:mnist_stage2_testset}.
First, we sample one to four digits from the original dataset, and randomly place them in four quadrants of the $64 \times 64$ space.
Note that the original samples have the $28 \times 28$ resolution, so we pad every sample with 2 pixels.
Then, we created image captions with syntactic variations.
To perform zero-shot-like generation, we applied different constraints between the train and test set, as shown in \Cref{fig:dataset_constraint}.
For example, white 0 on the upper left does not exist in the train set, while all 0's in the test set are white and placed in upper left. 
In total, we generated 300K samples for train, 3,000 for validation, and 3,000 for test set.

\noindent \textbf{Experiment Setting.}
Since aliasing occurs during downsampling, we train image quantizers with four different downsampling methods as baselines, including with Blurpool~\cite{zhang2019making}, as blurring before subsampling might be sufficient to anti-alias:
\texttt{StridedConv} (SC), \texttt{ConvBlurpool} (CB), \texttt{MaxPool} (MP), and \texttt{MaxBlurpool} (MB).
Then in Stage 2, similar to DALL-E, The decoder-only Transformer is trained by next-token prediction changing the quantizers pretrained in Stage 1. 
We measure the performance of a generator when TE-VQGAN (Ours) is used and vanilla VQGANs with various downsampling methods are used.
Experiments in \Cref{sec:exp} were conducted three times with random initialization.

\begin{table*}[ht]
\footnotesize
\caption{Semantic accuracy of text-to-image and image-to-text generation.}
\centering
\resizebox{\linewidth}{!}{
\begin{tabular}{lcccccccc}
\hline
              & \multicolumn{4}{c}{Text-to-Image}                                     & \multicolumn{4}{c}{Image-to-Text}                                    \\ \cmidrule(lr){2-5} \cmidrule(lr){6-9}
Model         & Quad1         & Quad2         & Quad3         & Quad4         & Quad1         & Quad2         & Quad3         & Quad4         \\ \hline
MB            & 0.724 (0.033) & 0.697 (0.066) & 0.635 (0.067) & 0.620 (0.077) & 0.587 (0.032) & 0.507 (0.021) & 0.433 (0.013) & 0.378 (0.020) \\
MP            & 0.713 (0.007) & 0.703 (0.015) & 0.669 (0.017) & 0.656 (0.028) & 0.561 (0.042) & 0.500 (0.038) & 0.412 (0.034) & 0.360 (0.039) \\
CB            & 0.753 (0.037) & 0.741 (0.034) & 0.682 (0.043) & 0.649 (0.055) & 0.563 (0.041) & 0.455 (0.056) & 0.362 (0.046) & 0.309 (0.043) \\
SC            & 0.812 (0.032) & 0.806 (0.030) & 0.790 (0.038) & 0.761 (0.051) & 0.786 (0.052) & 0.692 (0.063) & 0.595 (0.086) & 0.542 (0.106) \\
Ours & \textbf{0.931} (0.033) & \textbf{0.913} (0.032) & \textbf{0.890} (0.048) & \textbf{0.868} (0.059) & \textbf{0.825} (0.037) & \textbf{0.713} (0.066) & \textbf{0.621} (0.061) & \textbf{0.554} (0.065) \\ \hline
\end{tabular}
}
\label{tab:img_text_digit_acc}
\vspace{-3mm}
\end{table*}

\subsection{Text-to-Image Generation}
We measure the digit accuracy when the generated image has one (Quad1), two (Quad2), three (Quad3), and four digits (Quad4).
Each case has approximately 600 images.
As an example, in Quad2, given a text \texttt{`the lower right is blue nine, and the white two is on the upper left.'}, a rule-based caption parser will convert the text into \texttt{\{(LR, B, 9), (UL, W, 2)\}}.
With the parsed output, we crop the corresponding locations, lower right and upper left, from the generated image.
Then, we measure the accuracy of digits on these two cropped images using a pretrained classifier with a accuracy of 99.5\%.

We compare Ours, which uses the SC downsampling method with the orthogonality regularization, with four baselines.
As shown in \Cref{tab:img_text_digit_acc}, our method brings significant performance improvements compared to the baselines (MB, MP, CB, SC), clearly demonstrating the importance of translation equivariance in the quantized space.

\vspace{-2mm}
\subsection{Image-to-Text Generation}

Given a $64 \times 64$ image, the generator synthesizes a caption describing it. We measure its accuracy of digit identity using a simple rule-based parser.
As shown in \Cref{tab:img_text_digit_acc}, the gap between Ours and baselines in this task (I $\to$ T) is smaller than that of T $\to$ I, and the actual performance of I $\to$ T is significantly lower than that of T $\to$ I for all methods.
We conjecture that this difference comes from how texts and images are evaluated.
Since an image is classified as a whole, miss-predicting a code index or two might have little effect.
On the other hand, miss-predicting a specific text token, such as the digit token, can have a catastrophic impact.

\vspace{-2mm}
\subsection{Sample Efficiency}
\vspace{-3mm}

\begin{table}[ht]
\footnotesize
\centering
\caption{Digit accuracy of generated images in Quad1 varying the size of the train set.}
\begin{tabular}{lcc}
\hline
Size & VQGAN & TE-VQGAN (Ours) \\ 
\hline
300K          & 0.812 (0.032)                        & \textbf{0.931} (0.033)                    \\
100K          & 0.739 (0.037)                        & \textbf{0.919} (0.022)                    \\ \hline
Diff          & 0.073                        & 0.012                    \\ \hline
\end{tabular}
\label{tab:sample_efficiency}
\vspace{-2mm}
\end{table}

\noindent
We demonstrate that the translation-equivariant image quantization could improve the generator's ability to synthesize the shifted images, which are never shown at the training phase.
This is possible due to our model's consistent use of code indices for a given digit regardless of its position in the image.
From this result, we posit that the Stage 2 generator would be able to learn the relationship between image and text with an even smaller dataset. 
To verify this, we conduct an experiment where we train two generators on a small (100K) and a large (300K) training set.
We then measure the two generators' digit accuracy of T $\to$ I generation using the test set.

As we hypothesize, \Cref{tab:sample_efficiency} shows that the generator with TE-VQGAN is more robust to the reduced training set size than the one with vanilla VQGAN.
From this, we claim that translation-equivariance could be a potential solution to the scalability issue that modern generative models suffer from.


\section{Discussion}
\vspace{-2mm}

\begin{table}[ht]
\centering
\caption{Code usage, reconstruction loss, and perceptual loss.}
\resizebox{\linewidth}{!}{
\begin{tabular}{lcccccc}
\hline
\multirow{2}{*}{Dataset} & \multicolumn{2}{c}{Code usage} & \multicolumn{2}{c}{Recon loss} & \multicolumn{2}{c}{Perceptual loss} \\ 
\cmidrule(lr){2-3} \cmidrule(lr){4-5} \cmidrule(lr){6-7}
                         & VQGAN        & TE-VQGAN        & VQGAN         & TE-VQGAN       & VQGAN           & TE-VQGAN          \\ \hline
MNIST                    & 252          & 14              & 0.0079        & 0.0084         & 0.0036          & 0.0048            \\
FASHION                  & 250          & 18              & 0.0150        & 0.0187         & 0.0108          & 0.0198            \\ \hline
\end{tabular}
}
\label{tab:code_usage_and_recon_loss}
\vspace{-2mm}
\end{table}
\noindent
The first person to divide rainbow into 7 colors was Sir Isaac Newton. 
It is known that up to 207 colors of rainbows can be distinguished by human eye, but Newton expressed rainbows with only seven `essences'. 
Our methodology is similar in spirit: using the orthogonality regularization, only a few essence codes are used as described in \Cref{fig:idea}.

We count the number of codes that are used at least once in Stage 1.
Surprisingly, TE-VQGAN uses only 14 and 18 codes to represent colored digits and colored fashion items\footnote{We additionally conducted this experiment using Fashion-MNIST~\cite{fashionmnist}.}, as seen in \Cref{tab:code_usage_and_recon_loss}.
In other words, the 14 and 18 `essence' codes are sufficient to represent each dataset.

One possible limitation of using fewer essence codes is sacrifice in image fidelity.
Drawing a rainbow with only 7 colors will certainly miss its finer hue.
\Cref{tab:code_usage_and_recon_loss} shows the reconstruction loss $\left\| \mathbf{x}-\hat{\mathbf{x}} \right\|_1$ and the perceptual loss~\cite{zhang2018unreasonable} between $\mathbf{x}$ and $\hat{\mathbf{x}}$. 
This empirically confirms that, even though orthogonal regularization miss some visual information, its impact is marginal.
Since our ultimate aim lies in multimodal generation rather than reconstructing images, the ability to understand the semantics of the given condition is more important than reconstructing every bit of fine details. 
Therefore, considering the result from \Cref{sec:exp}, we can conclude that despite some minor drawbacks, we can achieve the more desirable goal of multimodal generation.


\section{Conclusion}

This work is a exploratory study that first explore \emph{translation equivariance in the image quantized space} and propose a simple yet effective way to achieve it.
Our proof-of-concept experiments demonstrate that our method improves image-text multimodal generation performance and sample efficiency.
Future research directions may include experimentation with real world datasets and diverse set of downstream tasks.

\clearpage 

\bibliographystyle{IEEEbib}
\bibliography{refs}

\end{document}